\documentclass[twocolumn]{autart}    % Enable this line and disable the 
\makeatletter
\def\ps@copyright{%
  \let\@oddhead\@empty
  \let\@evenhead\@empty
  \def\@oddfoot{\reset@font\hfil\thepage\hfil}%
  \let\@evenfoot\@oddfoot
}

\usepackage{graphicx}
\usepackage{color}

\usepackage{cite}
\usepackage{algcompatible, algorithm}
\usepackage{hyperref}
\usepackage{textcomp}

\usepackage{graphics} % for pdf, bitmapped graphics files
\usepackage{epsfig} % for postscript graphics files
\usepackage{subcaption} % for subfigure
\usepackage{amsmath} % assumes amsmath package installed
\usepackage{amssymb}  % assumes amsmath package installed
\usepackage[normalem]{ulem}
\usepackage{mathtools} % for vcentcolon
\usepackage{stmaryrd}
\usepackage{enumitem}
\theoremstyle{plain}
\newtheorem{theorem}{Theorem}[section] %[section]
\newtheorem{proposition}[theorem]{Proposition}
\newtheorem{lemma}[theorem]{Lemma}
\newtheorem{corollary}[theorem]{Corollary}
\theoremstyle{definition}

\theoremstyle{remark}
\newtheorem{remark}{Remark}[section]

\DeclareMathOperator*{\argmin}{arg\,min}

\newcommand{\jdiff}[1]{{#1}}

\begin{document}

\begin{frontmatter}
\title{ORFit: One-Pass Learning via Bridging Orthogonal Gradient Descent and Recursive Least Squares} % \thanksref{footnoteinfo}

\thanks[]{This paper was presented in part at the IEEE Conference on Decision and Control, Singapore, December 2022~\cite{min2022one} and at the American Control Conference, Canada, July 2024~\cite{cho2024pi}.}

\author[Cambridge]{Youngjae Min}\ead{yjm@mit.edu},  
\author[Cranfield]{Namhoon Cho}\ead{n.cho@cranfield.ac.uk},
\author[Cambridge]{Navid Azizan\corauthref{corrauth}}\ead{azizan@mit.edu}

\corauth[corrauth]{Corresponding author: N. Azizan. Tel. +1-617-715-4273.}

\address[Cambridge]{Laboratory for Information and Decision Systems, Massachusetts Institute of Technology, Cambridge, MA 02139, USA} 
\address[Cranfield]{Centre for Assured and Connected Autonomy, Cranfield University, Cranfield, Bedfordshire MK43 0AL, UK}


\begin{abstract}
While \jdiff{large machine learning models have shown remarkable} performance in various domains, their training typically requires iterating for many passes over the \jdiff{training data}. However, due to computational and memory constraints and potential privacy concerns, storing and accessing all the data is impractical in many real-world scenarios where the data arrives in a stream. In this paper, we investigate the problem of \emph{one-pass learning}, in which a model is trained on sequentially arriving data without retraining on previous datapoints. Motivated by \jdiff{the demonstrated effectiveness of overparameterized models and the phenomenon of benign overfitting, we propose} \emph{Orthogonal Recursive Fitting} ({\sf ORFit}), an algorithm for one-pass learning which seeks to perfectly fit each new datapoint while \jdiff{minimally altering} the predictions on previous datapoints. \jdiff{{\sf ORFit} updates the parameters in a direction orthogonal to past gradients, similar to \emph{orthogonal gradient descent} (OGD) in continual learning. We show that, interestingly, {\sf ORFit}'s update leads to an operation similar to the \emph{recursive least-squares} (RLS) algorithm in adaptive filtering but with significantly improved memory and computational efficiency, i.e., \emph{linear}, instead of quadratic, in the number of parameters. To further reduce memory usage, we leverage} the structure of the streaming data via an incremental principal component analysis (IPCA).
\jdiff{We show that using the principal components is minimax optimal, \emph{i.e.}, it minimizes the worst-case forgetting of previous predictions for unknown future updates.}
Further, we \jdiff{prove} that, for overparameterized linear models, the parameter vector obtained by {\sf ORFit} \jdiff{matches what the standard multi-pass stochastic gradient descent (SGD)} would converge to. %, \emph{i.e.}, it converges to the optimal parameter with the closest Euclidean distance to the initialization.
Finally, we extend our results to the nonlinear setting for highly overparameterized models, relevant for deep learning. Experimental results validate the effectiveness of the proposed method compared to the baselines.
\end{abstract}

\end{frontmatter}

\section{Introduction}

While \jdiff{large machine learning models have} been successful in numerous domains, their training is computationally demanding and requires iterating over the entire dataset multiple times. This hinders their deployment in many real-world settings such as robot learning, autonomy, and online decision making, where new datapoints are collected over time or become available sequentially. In such settings, storing all the datapoints and retraining the model at every step on all the data is extremely costly and often not feasible. In addition, in certain applications, storing the data may be prohibited for privacy reasons.
%However, making multiple passes over the dataset is unsuitable for many real-world applications: oftentimes, data arrives in a stream, and it is not practical to store entire data and revisit them multiple times with limited memory and computational resources. Also, storing data could be prohibited for privacy reasons when they include personal or confidential information.

Thus, it is very desirable to come up with algorithms that can learn incrementally or in an online fashion, rather than by \jdiff{repeatedly} iterating over the entire data many times. However, it is well-known that deep neural networks are prone to \jdiff{significantly} forgetting past information while learning new data, which is an issue referred to as \emph{catastrophic forgetting}~\cite{kemker2018measuring}.
This begs the question:
\begin{center}
    \emph{Can we learn streaming data efficiently without forgetting or retraining on previous data?}
\end{center}

This is a setting often referred to as \emph{one-pass learning}. 
More specifically, one-pass learning concerns the setting where \jdiff{({\romannumeral 1})} the algorithm makes an update \jdiff{using} the current datapoint without direct access to previous data; ({\romannumeral 2}) the new updates do not significantly affect the predictions on the previous data; and ({\romannumeral 3}) the computational and memory costs of each update must not grow with the iteration count.

\jdiff{One-pass learning and its variants have received increasing attention, and several studies have attempted to address them in various contexts~\cite{gao2013one,zhou2016one,ozyildirim2016one,hou2018one,zhao2021distribution}. For instance,} \cite{hu2021one} studied learning the ImageNet dataset in a single pass by revisiting some ``important'' previous datapoints at each learning step, and
\cite{wu2019large} investigated learning incremental ``batches'' on a large scale by correcting the classifier's bias towards new data. 
However, both methods train on previous data and are not adequate for one-pass learning.
On the other hand, \cite{rai2009streamed} proposed an effective one-pass learning method for support vector machines. However, their method is tailored to the specific setting of support vector machines.
Further, \cite{sahoo2018online} proposed a one-pass deep learning algorithm, but it relies on a specific network architecture and is vulnerable to forgetting the previous data unless the data is consistently arriving from the same distribution.

One-pass learning is also closely related to a classical problem studied in the context of control and estimation theory.
More specifically, a classical algorithm known as recursive least-squares (RLS) (see, \emph{e.g.}, \cite{sayed2003fundamentals}) tackles one-pass learning for linear models (as elaborated in Section~\ref{sec:rls}).
However, there are two limitations of the standard RLS: ({\romannumeral 1}) it suffers from high computational and memory costs, and ({\romannumeral 2}) it is not well-suited for the overparameterized setting where zero training loss is desired.
The main focus of this work is to develop a method that overcomes these limitations.
Related to the overparameterized setting, a few works have recently discussed utilizing RLS to train popular deep neural networks such as FNN, CNN, RNN, and LSTM~\cite{zhang2021revisiting,yu2022recursive}.
However, these works are empirical in nature and consider multi-pass learning with mini-batches.
Moreover, the theoretical properties of RLS for one-pass learning are not studied in those works.

\subsection{Contributions}
Our main contributions can be summarized as follows.
\begin{itemize}
    \item  We develop Orthogonal Recursive Fitting ({\sf ORFit}), an algorithm for one-pass learning in the overparameterized setting, which fits new data on the fly while updating the parameters in a direction that causes the least change to the predictions on previous data.
    Our algorithm uses memory efficiently by exploiting the structure of the streaming data via \jdiff{an} incremental principal component analysis (IPCA) to extract the essential information for the update.
    \jdiff{We further generalize our algorithm to learn from batches of data, which can be viewed as a continual learning method.}
    (\S\ref{sec:orfit})
    \item Through the proposed method, we establish an interesting connection between two different algorithms from adaptive filtering and machine learning, namely, the recursive least-squares (RLS) algorithm and the orthogonal gradient descent (OGD). \jdiff{Our method updates the parameters in a direction orthogonal to past gradients, thereby ensuring minimal disruption of previous predictions, similar to OGD. We show that this update leads to an operation similar to the RLS algorithm, but with significantly improved memory and computational cost. (\S\ref{sec:theory})}
    
    \item
    \jdiff{We characterize the behavior of the proposed method in the overparameterized feature-based linear setting. We prove that {\sf ORFit}, in one pass, finds the same parameter vector that the standard multi-pass SGD would asymptotically converge to. Additionally, we show that using the principal components to reduce memory usage is minimax optimal in terms of forgetting.} (\S\ref{sec:theory})
    
    \item  We demonstrate the practicality of our approach and corroborate our theoretical findings through various experiments. (\S\ref{sec:exp})
    
    \item We discuss extensions of our results to overparameterized nonlinear models, relevant for deep learning. (\S\ref{sec:deep})
\end{itemize}

\jdiff{
\subsection{Connections to Related Notions}
One-pass learning shares some similarities with other settings that deal with streaming data such as \emph{online learning} and \emph{incremental learning}. 
While these terms are often inconsistently defined in the literature, one may distinguish them from one-pass learning based on whether we learn a single datapoint or a batch of data at a time~\cite{nallaperuma2019online}. Additionally, although both of these approaches learn from streaming data, they often make assumptions about the distribution of the data. 
Compared to these settings, one-pass learning requires explicit efforts to not alter the predictions on the previous data while learning new data. Thus, it aims to preserve the predictions even when the new data comes from a completely different distribution. Another related setting is \emph{continual learning} where batches of data from different tasks are sequentially learned~\cite{delange2021continual}. 
Note that a one-pass learning algorithm is applicable to this setting to learn multiple tasks without the knowledge of the boundaries between the tasks. 
}

\section{Preliminaries}

\subsection{One-Pass Learning}

Let \jdiff{$f(x;w)\in\mathbb{R}^c$} be a model that the agent is trying to fit, where  $x\in\mathcal{X}\subset\mathbb{R}^d$ is the input and $w\in\mathbb{R}^p$ is the parameter (weight) vector.
Consider a sequentially arriving stream of data $\left((x_{k}, y_{k})\right)_{k=1}^{K}$ where  $x_{k}\in\mathcal{X}$ and \jdiff{$y_{k}\in\mathcal{Y}\subset\mathbb{R}^c$}. In overparameterized models, we have $p\geq K$ (and often $p\gg K$). For a loss function $\ell(\cdot,\cdot)$, let $f_{k}(w)\vcentcolon=f(x_{k};w)$ and $\ell_{k}(w)\vcentcolon=\ell(y_{k}, f_{k}(w))$. Then, one-pass learning considers the setting where, given an initial parameter $w_0\in\mathbb{R}^p$, it updates the parameter $w_i\in\mathbb{R}^p$ after the new data $(x_{i}, y_{i})$ arrived without revisiting previous datapoints $\{(x_{k}, y_{k})\}_{k=1}^{i-1}$.

For concreteness, we first focus on \jdiff{an overparameterized feature-based linear model $f(x;w)=\Phi(x)^\top w=[\phi_1(x) \cdots \phi_c(x)]^\top w$ with feature maps $\phi_j:\mathcal{X}\rightarrow\mathbb{R}^{p}$ for $j\in[c]:=\{1,\dots,c\}$}. We then discuss the extension of the results to nonlinear models (\emph{e.g.,} overparameterized architectures in deep learning) in Section~\ref{sec:deep}.

Before introducing our algorithm and the results, we briefly review two related algorithms capable of one-pass learning, proposed in two different literatures.

\subsection{Recursive Least-Squares}
\label{sec:rls}
First, we briefly review recursive least-squares (RLS) from the control/estimation theory literature (refer to~\cite{sayed2003fundamentals} for details). \jdiff{In its standard form, RLS considers a linear model $f(x;w) = w^\top x\in\mathbb{R}$ for a stream of data $\left((x_k, y_k)\right)_{k=1}^K$ with $x\in\mathbb{R}^p$, $y_k\in\mathbb{R}$.} At every step $i$, it aims to find a parameter vector that solves the following regularized least-squares problem:
\begin{equation}    \label{eq:ls}
    w^{(RLS)}_i = \arg\min_w \sum_{k=1}^{i} (y_k - w^\top x_k)^2 + \lVert w-w_0 \rVert^2_\Pi,
\end{equation}
where $\lVert x \rVert_\Pi\vcentcolon=\sqrt{x^\top\Pi x}$ for a $p\times p$ positive-definite matrix $\Pi$ and $w_0$ is an initial parameter estimate.
Note that the system is underdetermined/overparameterized, and the regularization term in~\eqref{eq:ls} is necessary for the solution to be uniquely defined.
While there is a closed-form solution for \eqref{eq:ls} given by $w^{(RLS)}_i = (\Pi + X_i^\top X_i)^{-1}(X_i^\top Y_i+\Pi w_0)$ with $X_i = [x_1\;x_2 \dots \;x_i]^\top$ and $Y_i = [y_1\;y_2 \dots \;y_i]^\top$,  computing the solution directly requires storing all the previous data as well as recomputing the inverse of the covariance matrix for every new datapoint.
RLS bypasses this issue by computing the new solution $w^{(RLS)}_i$ of \eqref{eq:ls} recursively from $w^{(RLS)}_{i-1}$ and $(x_i,y_i)$.

We elaborate the algorithm more formally for a general version of RLS called \emph{exponentially weighted recursive least-squares} (EW-RLS)~\cite{sayed2003fundamentals}. Consider the following problem:
\begin{equation} \label{eq:weighted_ls}
    w^{(RLS)}_i = \arg\min_w \sum_{k=1}^{i} \lambda^{i-k}(y_k - w^\top x_k)^2 + \lambda^i \lVert w-w_0 \rVert^2_\Pi,
\end{equation}
with a forgetting factor $0<\lambda\leq 1$. Note that this reduces to the problem of the vanilla RLS~\eqref{eq:ls} when $\lambda = 1$. The exact solution of it is recursively updated as follows:
\begin{equation}
\begin{cases}
    w^{(RLS)}_i = w^{(RLS)}_{i-1} + \dfrac{P_{i-1} x_i}{\lambda^i + x_i^\top P_{i-1} x_i} (y_i - x_i^\top w^{(RLS)}_{i-1})\,,\\
    P_i = P_{i-1} - \dfrac{P_{i-1} x_i x_i^\top P_{i-1}}{\lambda^i + x_i^\top P_{i-1} x_i},
\end{cases}
\label{eq:rls_update}
\end{equation}
with $w^{(RLS)}_0=w_0$ and $P_0 = \Pi^{-1}$. Here, $P_i$ can be alternatively written as $P_i = [\Pi + X_i^\top \Lambda_i X_i]^{-1}$ where $\Lambda_i = \mathrm{diag}(\lambda^{-1}, \lambda^{-2}, \dots, \lambda^{-i})$.
\jdiff{Note that the update rule~\eqref{eq:rls_update} of EW-RLS can be highly inefficient as its memory and computational complexities are $O(p^2)$, which is prohibitive for one-pass learning in overparameterized models, where we often have $p\gg K$. In contrast, we develop an algorithm with linear complexities (see Section~\ref{subsec:connection} for more details).}

\subsection{Orthogonal Gradient Descent}
An algorithm called orthogonal gradient descent (OGD)~\cite{farajtabar2020orthogonal} has been proposed in the context of machine learning for a different but related problem. 
More specifically, \cite{farajtabar2020orthogonal} considers a \emph{continual learning} setting in which tasks $\{T_1, T_2, \dots \}$ arrive sequentially, and each task consists of a set of datapoints.
Continual learning can be understood as a ``batch'' version of one-pass learning.
At a high level, when the $i$-th task $T_i$ arrives, OGD updates the parameter \jdiff{using} new samples from $T_i$ in a way that causes minimal changes to the predictions for previous tasks $\{T_b\}_{b=1}^{i-1}$.
The gradient of the model $f(x;w)\in\mathbb{R}$ on a datapoint $x_j$ with respect to the parameter, $\nabla_w f(x_j;w)$, is the direction in the parameter space that causes the most change to the prediction on that datapoint. Thus, moving orthogonal to this direction, locally, keeps the prediction unchanged, which is the main idea behind OGD.
More formally, the update direction is computed via projecting the current \jdiff{gradient}\footnote{\jdiff{To be precise, the update rule in OGD uses $g$ to denote the gradient of the loss, rather than the gradient of the model; however, as long as the former is nonzero, their direction is the same.}} \jdiff{$g$} onto the subspace orthogonal to $\mathcal{G}\vcentcolon=\text{span}\{\bigcup_{b=1}^{i-1}\{\nabla_w f(x;w_b)\}_{(x,y)\in T_b}\}$:
\begin{equation}
    \tilde{g} = g - \sum_{v\in S} \textup{proj}_{v} (g),
\label{eq:ogd_projection}
\end{equation}
where $S$ is an orthogonal basis for $\mathcal{G}$ and $\textup{proj}_{v}(g) \vcentcolon= (g^\top v/\lVert v \rVert^2) v = (vv^\top/\lVert v \rVert^2) g$. The orthogonal basis $S$ is incrementally updated through the Gram-Schmidt procedure.

\begin{figure}[t]
	\centering
    \includegraphics[width=.4\textwidth]{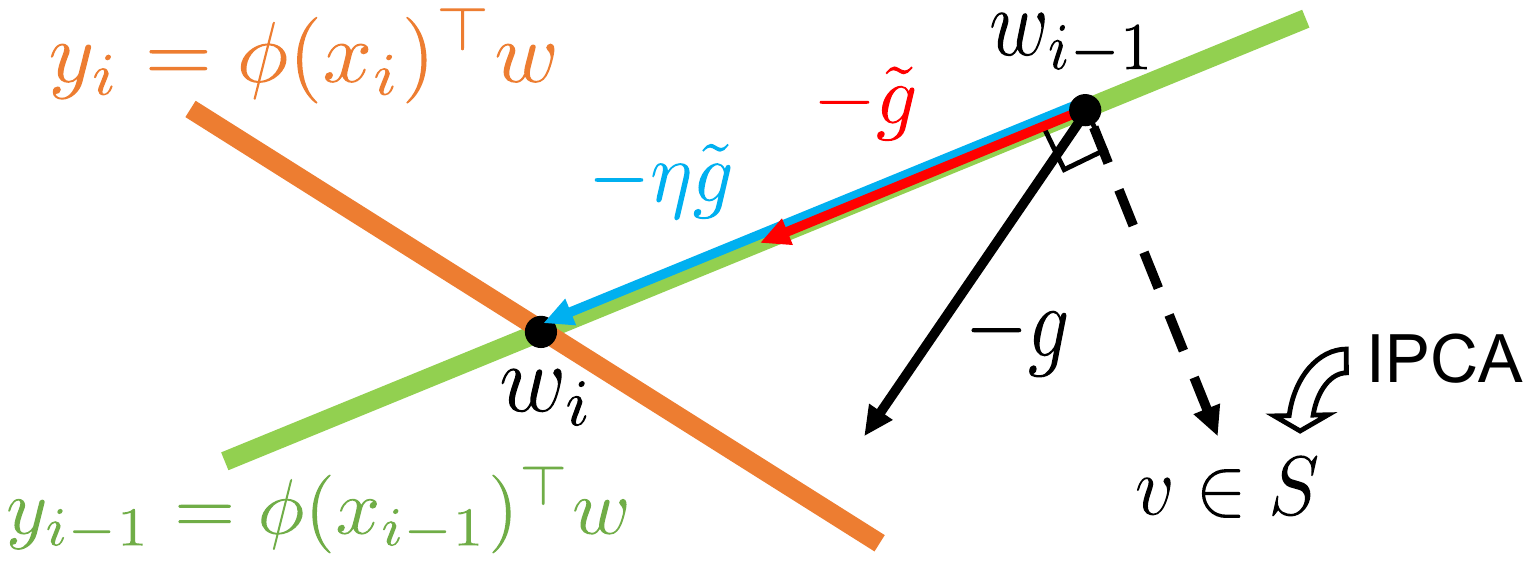}
	\caption{An illustration of {\sf ORFit} in the parameter space for a linear model. The parameter $w_{i-1}$ fits the previous datapoints $\{(x_k,y_k)\}_{k=1}^{i-1}$. The set $S$ (which is updated incrementally) consists of the directions moving towards which causes the most change in the predictions on previous data, and thus, moving orthogonal to $S$ keeps the predictions intact. Given a new datapoint $(x_i,y_i)$, projecting its corresponding gradient $g$ to the orthogonal complement of the subspace spanned by $S$ yields the new update direction $\tilde{g}$. {\sf ORFit} finds a new parameter $w_i$ along the direction of $-\tilde{g}$ which fits the new datapoint $(x_i, y_i)$ within a single step, while still fitting the previous data $\{(x_k,y_k)\}_{k=1}^{i-1}$.}
	\label{fig:algorithm}
\end{figure}

\section{Orthogonal Recursive Fitting} \label{sec:orfit}

In this section, we propose a one-pass learning algorithm called \emph{orthogonal recursive fitting} ({\sf ORFit}). 
The algorithm consists of three main components: ({\romannumeral 1}) orthogonal update of the parameter motivated by OGD; ({\romannumeral 2}) interpolation (perfect fitting) of new data in a single step; and ({\romannumeral 3}) efficient use of memory via incremental summary. 
We describe the details of each component in what follows, starting from (i) and (ii). See Fig.~\ref{fig:algorithm} for an illustration.

\subsection{Orthogonal Recursive Update}\label{sec:orfit_infmem}
We start by considering OGD directly applied to the one-pass learning setting \jdiff{with a scalar-output model $f(x;w)\in\mathbb{R}$, \emph{i.e.}, $c=1$}. By treating each task $T_k$ in the continual learning setting as consisting of a single datapoint, OGD will run multiple gradient descent steps on the single datapoint. 
While perfect fitting/interpolation is often desired in highly overparameterized models~\cite{belkin2019reconciling,bartlett2020benign}, it will take many iterations for OGD to perfectly fit the datapoint.
Instead, inspired by the trends in meta-learning, we consider a one-step learning scheme that only runs a \emph{single} gradient step to interpolate the new datapoint.
We first begin with the following result that serves as a building block for our algorithm design; see Appendix~\ref{app:proof_step} for a proof.
\begin{lemma} \label{lem:step}
    Consider a \jdiff{feature-based linear model $f(x;w)=\phi(x)^\top w\in\mathbb{R}$}, and let $\tilde{g}$ be the projection defined by \eqref{eq:ogd_projection} of any vector $g\in\mathbb{R}^p$. Then, for any step size $\eta\in\mathbb{R}$, the new parameter $w' = w - \eta \tilde{g}$ preserves the predictions on the previous datapoints, \emph{i.e.}, $f(x;w') = f(x;w)$ for all $(x,y)\in\bigcup_{k=1}^{i-1}T_k$.
\end{lemma}

The main takeaway of Lemma~\ref{lem:step} is that the predictions for the previous datapoints do not change when we update the model along the direction $\tilde g$. Hence, we may choose $\eta$ so that the updated parameter $w'$ can perfectly fit the new datapoint, say $(x', y')$, as \jdiff{$y' = \phi(x')^\top w' = \phi(x')^\top (w - \eta \tilde{g})$}. 
Following this principle, a straightforward calculation yields the following update rule:
\begin{equation} \label{eq:orfit1_infmem}
\jdiff{
\begin{cases}
    \tilde{g}_{i-1} = \nabla f_{i}(w_{i-1}) - \sum_{v\in S_{i-1}} \textup{proj}_{v} (\nabla f_{i}(w_{i-1}))\,,\\
    w_i = w_{i-1} - \eta_{i-1} \tilde{g}_{i-1}\,,\\
    S_i = S_{i-1} \bigcup \{\tilde{g}_{i-1}\},
\end{cases}}
\end{equation}
for $i\geq 1$ where $S_0$ is the empty set, $w_0$ is the initial weight vector, and the optimal step size is chosen as
\begin{equation} \label{eq:stepsize}
    \eta_{i-1} = \frac{1}{\nabla f_i(w_{i-1})^\top  \tilde{g}_{i-1}} (f_i(w_{i-1}) - y_i),
\end{equation}
\jdiff{assuming $\tilde{g}_{i-1}\neq0$.}
See Algorithm~\ref{alg:orfit1_infmem} for the detailed procedure. \jdiff{Compared to OGD, the basis is kept orthonormal for more efficient projection.}
For intuition, we note that the optimal step size~\eqref{eq:stepsize} is typically small for highly overparameterized models; there are many parameter vectors in the vicinity of the current solution that perfectly fit the new datapoint~\cite{azizan2018stochastic,li2018learning,allen2019convergence}.

\begin{algorithm}[!t]
\caption{Orthogonal Recursive Fitting ({\sf ORFit}) for scalar-output model ($c=1$) without memory restriction}
\label{alg:orfit1_infmem}
\begin{algorithmic}[1] 
\renewcommand{\algorithmicrequire}{\textbf{Input:}}
\renewcommand{\algorithmicensure}{\textbf{Output:}}
\jdiff{
\REQUIRE Data sequence $\left((x_k, y_k)\right)_{k=1}^{K}$
\ENSURE  The optimal parameter $w$
\STATE \textbf{Initialize} $U\leftarrow [\;\;], \; w\leftarrow w_0$
\FOR {$k = 1,2,\dots,K$}
\STATEx \(\triangleright\) Update parameter
\STATE $g \leftarrow$ Sample model gradient $\nabla f_k(w)$
\STATE $\tilde{g} \leftarrow g - U(U^\top g)$
\STATE $\eta \leftarrow (f_k(w) - y_k)/(g^\top \tilde{g})$
\STATE $w \leftarrow w - \eta \tilde{g}$
\STATEx \(\triangleright\) Update orthonormal basis
\STATE $U \leftarrow [U \;\tilde{g}/\lVert \tilde{g} \rVert]$
\ENDFOR
}
\end{algorithmic} 
\end{algorithm}

\begin{remark}[Computational overhead]
    It is important to note that all the quantities appearing in \eqref{eq:stepsize} are typically available in the gradient-based optimization setting, and hence, there is no computational overhead for computing the stepsize \eqref{eq:stepsize}.
\end{remark}
\begin{remark}[Nonlinear models]
    Although the update rule \eqref{eq:orfit1_infmem} is derived based on linear models, it can also be applied to highly overparameterized \emph{nonlinear} models such as deep neural networks, as we will discuss in Section~\ref{sec:deep}.
\end{remark}
Another distinction between the update rule~\eqref{eq:orfit1_infmem} and OGD lies in the update of the orthogonal basis $S_i$: OGD utilizes ``fresher'' gradient at the updated parameter $w_i$ by $S_i = S_{i-1} \bigcup \{\nabla f_{i}(w_{i}) - \sum_{v\in S_{i-1}} \textup{proj}_{v} (\nabla f_{i}(w_{i}))\}$. 
Although the two bases actually span the same subspace for linear models, it turns out that {\sf ORFit} in~\eqref{eq:orfit1_infmem} leads to a natural generalization for nonlinear models; see Section~\ref{sec:deep} for details.

\subsection{Extension to Vector-Output Model}
\jdiff{
Now, we extend the update rule~\eqref{eq:orfit1_infmem} to learn a general vector-output model $f(x;w)\in\mathbb{R}^c$, which subsumes the scalar-output case ($c=1$).
\begin{remark}[Other forms of vector-output model]
    Note that we could also consider a vector-output model $f(x;w_1,\dots,w_c)=[\phi_1(x)^\top w_1 \cdots \phi_c(x)^\top w_c]^\top$ which has a separate parameter vector $w_j\in\mathbb{R}^p$ for each output dimension $j\in[c]$. Especially, with a shared feature vector $\phi(x)=\phi_1(x)=\cdots=\phi_c(x)\in\mathbb{R}^p$, the model could be $f(x;w_1,\dots,w_c)=[w_1 \cdots w_c]^\top \phi(x)$. An example is learning with a pre-trained neural network model by fine-tuning only the last fully connected layer while freezing the earlier layers. In this case, the feature vector $\phi(x)$ corresponds to the output of the penultimate layer. However, such models could be dealt with as multiple separate scalar-output models, one for each output dimension, with their own parameters. Then, {\sf ORFit} for $c=1$ is enough to update them independently.
\end{remark}
}

\jdiff{
As in the update rule~\eqref{eq:orfit1_infmem}, we learn the vector-output model $f(x;w)=\Phi(x)^\top w\in\mathbb{R}^c$ in one pass by making orthogonal updates of the parameter and interpolating new datapoint in a single step. At each iteration $i$, we first choose an update direction by projecting the model gradients onto the subspace orthogonal to the previous gradients. Unlike in the scalar-output case, we have multiple gradient directions, one for each output, in the form of a Jacobian matrix. Thus, we project each gradient direction onto the orthogonal subspace and find an update direction as a linear combination of the projected gradients. This update direction is still orthogonal to the previous gradients to preserve the predictions on previous data. 

Then, we choose the coefficients of the linear combination to interpolate the new datapoint, similarly to how we have computed the optimal step size in~\eqref{eq:stepsize}. Assuming that new model gradients are not spanned by the previous gradients, the extended update rule is 
}
\jdiff{
\begin{equation}
\label{eq:orfit2_infmem}
\begin{cases}
    G_{i-1} = 
    \left(\frac{\partial f_{i}(w_{i-1})}{\partial w}\right)^\top,\\
    \tilde{G}_{i-1} = G_{i-1} - \sum_{v\in S_{i-1}} \textup{proj}_{v} (G_{i-1})\,,\\
    w_i = w_{i-1} - \tilde{G}_{i-1} (G_{i-1}^\top \tilde{G}_{i-1})^{-1} (f_i(w_{i-1})-y_i)\,,\\
    S_i = S_{i-1} \bigcup \{\textup{orth}(\textup{col}(\tilde{G}_{i-1}))\},
\end{cases}
\end{equation}
where $\textup{proj}_{v}(G)$ is the column-wise projection of $G$ onto the direction of $v$, $\textup{col}(G)$ is the set of the columns of $G$, and $\textup{orth}(S)$ is an orthonormalization of $S$. This update rule reduces to \eqref{eq:orfit1_infmem} when $c=1$. The linearity of the model with respect to the parameters ensures that this update step fits the new datapoint. We theoretically analyze this update in detail in Section~\ref{subsec:solution}.
}

Although the update rule~\eqref{eq:orfit2_infmem} does not access previous datapoints, they still require storing the orthogonal basis $S_i$, whose size grows linearly in the number of visited datapoints.
This is not desirable in practice when one needs to train the model on a large dataset. 
We address this issue next.

\subsection{Incremental Summary of Memory}
\label{sec:orfit_ipca}
In this section, we overcome the aforementioned memory issue by utilizing the structure of the streaming dataset.
The main idea is to \jdiff{approximate the orthogonal basis $S$ in a lower dimension} using an incremental principal component analysis (IPCA) algorithm, known as the sequential Karhunen–Loeve (SKL) algorithm proposed in~\cite{levey2000sequential}.
IPCA is a memory-efficient variant of PCA  that enables sequential update for streaming/large datasets. 
Let us formally describe how \jdiff{we utilize IPCA to incrementally approximate} the orthogonal basis.

Consider \jdiff{the past model gradients $\cup_{k=1}^i \textup{col}(G_i)$ spanned by the orthogonal basis $S$ as in~\eqref{eq:orfit2_infmem}.} Let  the singular value decomposition (SVD) of \jdiff{$A=[G_1\; G_2\,\dots G_i]$} be  $A=U\Sigma V^\top$. 
Here, the crucial information of the SVD is the left-singular vectors $\mathrm{col}(U)$ that form an orthonormal basis for $\mathrm{span}(S)$.
This information can be used to come up with a rank-$m$ approximation of \jdiff{$S$} by using the principal components corresponding to the top $m$ singular values. \jdiff{As will be shown, this choice of approximation is minimax optimal, \emph{i.e.}, it minimizes the worst-case forgetting of previous predictions for unknown future updates. We will formally establish this in Section~\ref{sec:pca_theory}.}

Now suppose that the orthogonal basis $S$ is augmented \jdiff{with the new gradients $G$ as in~\eqref{eq:orfit2_infmem}. We first want to efficiently update $U$ for the new basis instead of recomputing the SVD from scratch. The orthogonal components of $G$ not spanned by $S$ can be expressed as $\tilde{G}=G-U(U^\top G)$. Letting $G_\text{orth}$ denote the column-wise orthonormalization of $\tilde{G}$, the new basis matrix can be represented as:
\begin{align}
    \begin{bmatrix}
        A & G
    \end{bmatrix}
    &=
    \begin{bmatrix}
        U\Sigma V^\top
        & (UU^\top + G_\textup{orth} G_\textup{orth}^\top) G
    \end{bmatrix}\\
    &=
    \begin{bmatrix}
        U & G_\textup{orth}
    \end{bmatrix}
    \begin{bmatrix}
        \Sigma & U^\top G\\
        0 & G_\textup{orth}^\top G
    \end{bmatrix}
    \begin{bmatrix}
        V^\top & 0\\
        0 & I
    \end{bmatrix}
    \\
    &= \left(
    \begin{bmatrix}
        U & G_\textup{orth}
    \end{bmatrix} \tilde{U}
    \right) \tilde{\Sigma}
    \left( \tilde{V}^\top
    \begin{bmatrix}
        V^\top & 0\\
        0 & I
    \end{bmatrix}
    \right), \label{eq:svd_update}
\end{align}
where $\tilde{U}\tilde{\Sigma}\tilde{V}^\top$ is the SVD of $\begin{bmatrix}\Sigma & U^\top G\\ 0 & G_\textup{orth}^\top G \end{bmatrix}$.}
Then, \eqref{eq:svd_update} is the SVD of the new basis matrix. 
Hence, to update $U$, one can directly use the information from the previous iteration, namely $U$ and $\Sigma$. 
The important aspect here is that the update can be made \emph{without storing} $V$ and \emph{without having to recompute} the SVD of the new \jdiff{gradient} matrix. Finally, one can store only the top $m$ singular values in $\Sigma$ and their corresponding components in $U$.
By repeatedly applying this IPCA algorithm in addition to~\jdiff{\eqref{eq:orfit2_infmem}}, we obtain \emph{orthogonal recursive fitting ({\sf ORFit})}. See Algorithm~\ref{alg:orfit_ipca} for the detailed procedure.

% \begin{algorithm}[!t]
% \caption{Orthogonal Recursive Fitting ({\sf ORFit}) for scalar-output model ($c=1$) with memory limit $m$}
% \begin{algorithmic}[1] \label{alg:orfit1_ipca}
% \renewcommand{\algorithmicrequire}{\textbf{Input:}}
% \renewcommand{\algorithmicensure}{\textbf{Output:}}
% \jdiff{
% \REQUIRE Data sequence $\left((x_k, y_k)\right)_{k=1}^{K}$
% \ENSURE  The optimal parameter $w$
% \STATE \textbf{Initialize} $U\leftarrow [\;\;], \;\Sigma \leftarrow [\;\;], \; w\leftarrow w_0$
% \FOR {$k = 1,2,3,\dots$}
% \STATE \(\triangleright\) Update parameter
% \STATE $g \leftarrow$ Sample model gradient $\nabla f_k(w)$
% \STATE $\tilde{g} \leftarrow g - U (U^\top g)$
% \STATE $\eta \leftarrow (f_k(w) - y_k)/(g^\top  \tilde{g})$
% \STATE $w \leftarrow w - \eta \tilde{g}$
% \STATE \(\triangleright\) Update basis
% \STATE $u \leftarrow g/\lVert g \rVert$
% % \STATE $R \leftarrow \begin{bmatrix} \Sigma & U^\top  \tilde{g}\\ 0 & u^\top \tilde{g}\end{bmatrix}$
% \STATE $\Tilde{U}, \Sigma \leftarrow$ Compute SVD of $\begin{bmatrix} \Sigma & U^\top g'\\ 0 & u^\top g\end{bmatrix}$ % = \Tilde{U}\Tilde{\Sigma}\Tilde{V}^\top
% % \STATE $\Sigma \leftarrow \Tilde{\Sigma}$
% \STATE $U \leftarrow [U\; u]\Tilde{U}$
% \STATE $U, \Sigma \leftarrow$ top $m$ singular vectors/values in $U, \Sigma$
% \ENDFOR
% }
% \end{algorithmic} 
% \end{algorithm}

\begin{algorithm}[!t]
\caption{Orthogonal Recursive Fitting ({\sf ORFit}) for vector-output model with memory limit $m$}
\label{alg:orfit_ipca}
\begin{algorithmic}[1]
\renewcommand{\algorithmicrequire}{\textbf{Input:}}
\renewcommand{\algorithmicensure}{\textbf{Output:}}
\jdiff{
\REQUIRE Data sequence $\left((x_{k}, y_{k})\right)_{k=1}^{K}$
\ENSURE  The optimal parameter $w$
\STATE \textbf{Initialize} $U\leftarrow [\;\;],\; \Sigma\leftarrow[\;\;], \; w\leftarrow w_0$
\FOR {$k = 1,2,\dots,K$}
\STATEx \(\triangleright\) Update parameter
\STATE $G \leftarrow \left(\frac{\partial f_{k}(w)}{\partial w}\right)^\top$
\STATE $\tilde{G} \leftarrow G - U (U^\top G)$
\STATE $w \leftarrow w - \tilde{G}\big(G^\top \tilde{G}\;\big)^{-1}(f_k(w)-y_k)$
\STATEx \(\triangleright\) Update orthonormal basis
\STATE $G_{\mathrm{orth}} \leftarrow \mathrm{Orthogonalize } \;\tilde{G}$
% \STATE $R \leftarrow \begin{bmatrix} \Sigma & U^\top  \tilde{g}\\ 0 & u^\top \tilde{g}\end{bmatrix}$
\STATE $\Tilde{U}, \Sigma \leftarrow$ Compute SVD of $\begin{bmatrix} \Sigma & U^\top G\\ 0 & G_{\mathrm{orth}}^\top \tilde{G}\end{bmatrix}$ % = \Tilde{U}\Tilde{\Sigma}\Tilde{V}^\top
% \STATE $\Sigma \leftarrow \Tilde{\Sigma}$
\STATE $U \leftarrow [U\; G_{\mathrm{orth}}]\Tilde{U}$
\STATE $U, \Sigma \leftarrow$ top $m$ singular vectors/values in $U, \Sigma$
\ENDFOR
}
\end{algorithmic} 
\end{algorithm}

The main advantage of {\sf ORFit} is its  {\bf computational/memory efficiency}.
By only storing the top $m$ components, we can reduce the memory size \jdiff{at each iteration $i$ from $O(icp)$} to $O(mp)$. 
Moreover, the additional computational overhead to perform IPCA as well as the total time complexity of {\sf ORFit} at each step is \jdiff{$O((m+c)^2p)$, while $O(icp)$ is required without IPCA}.
Hence, {\sf ORFit} can reduce both the computation and the memory complexity\jdiff{, especially for overparameterized models with large $p$, by appropriately choosing $m$}.

\subsection{{\sf Batch-ORFit}: Batch Update and Continual Learning}
\jdiff{
In addition, we further extend the algorithm to learn batches, \emph{i.e.}, chunks, of data, which can be considered as a continual learning method. As we will see, it turns out that batch learning can be accomplished in the same manner as the vector-output model case.
Let $f(x;w)\in\mathbb{R}^c$ be a model that we want to fit, where  $x\in\mathcal{X}\subset\mathbb{R}^d$ is the input and $w\in\mathbb{R}^p$ is the parameter (weight) vector.
Consider a sequentially arriving stream of batched data $(\{(x_{b,k}, y_{b,k})\}_{k=1}^{n_b})_{b=1}^{B}$ where  $x_{b,k}\in\mathcal{X}$ and $y_{b,k}\in\mathcal{Y}\subset\mathbb{R}^c$. In overparameterized models, we have $p\geq \sum_{b=1}^B n_b$ (and often $p\gg \sum_{b=1}^B n_b$). For a loss function $\ell(\cdot,\cdot)$, let $f_{b,k}(w)\vcentcolon=f(x_{b,k};w)$ and $\ell_{b,k}(w)\vcentcolon=\ell(y_{b,k}, f_{b,k}(w))$. Then, the batch/continual learning problem considers the setting where, given an initial parameter $w_0\in\mathbb{R}^p$, we update the parameter $w_i\in\mathbb{R}^p$ after the new data batch $\{(x_{i,k}, y_{i,k})\}_{k=1}^{n_i}$ arrived without revisiting previous datapoints $\left\{\{(x_{b,k}, y_{b,k})\}_{k=1}^{n_b}\right\}_{b=1}^{i-1}$.
}

\jdiff{
The new data batch can be learned as if we are fitting an augmented single datapoint:
\begin{equation}
    \underbrace{[y_{i,1}; \cdots; y_{i,n_i}]}_{=:Y_i} =
    \underbrace{[f_{i,1}(w); \cdots; f_{i,n_i}(w)]}_{=:F_i(w)},
\end{equation}
where $[(\cdot) ; (\cdot)]$ denotes row-wise concatenation. Note that the batch size $n_i$ affects the effective output dimension of $Y_i$ and $F_i$. Then, we can achieve this interpolation through the following update rule that generalizes \eqref{eq:orfit2_infmem}:
\begin{equation}
\label{eq:borfit_infmem}
\begin{cases}
    G_{i-1} = \Big[
    \left(\frac{\partial f_{i,1}(w_{i-1})}{\partial w}\right)^\top \cdots 
    \left(\frac{\partial f_{i,n_i}(w_{i-1})}{\partial w}\right)^\top
    \Big],\\
    \tilde{G}_{i-1} = G_{i-1} - \sum_{v\in S_{i-1}} \textup{proj}_{v} (G_{i-1})\,,\\
    w_i = w_{i-1} - \tilde{G}_{i-1} (G_{i-1}^\top \tilde{G}_{i-1})^{-1} (F_i(w_{i-1}) - Y_i)\,,\\
    S_i = S_{i-1} \bigcup \{\textup{orth}(\textup{col}(\tilde{G}_{i-1}))\},
\end{cases}
\end{equation}
as presented in Algorithm~\ref{alg:borfit_infmem}. Since this method also faces the growing memory issue, we can utilize IPCA to accommodate memory constraints. See Algorithm~\ref{alg:borfit_ipca} for the detailed procedure.
}

\begin{algorithm}[!t]
\caption{{\sf Batch-ORFit} without memory restriction}
\label{alg:borfit_infmem}
\begin{algorithmic}[1]
\renewcommand{\algorithmicrequire}{\textbf{Input:}}
\renewcommand{\algorithmicensure}{\textbf{Output:}}
\jdiff{
\REQUIRE Data batch sequence $\left(\{(x_{b,k}, y_{b,k})\}_{k=1}^{n_b}\right)_{b=1}^{B}$
\ENSURE  The optimal parameter $w$
\STATE \textbf{Initialize} $U\leftarrow [\;\;], \; w\leftarrow w_0$
\FOR {$b = 1,2,\dots,B$}
\STATEx \(\triangleright\) Update parameter
\STATE $G \leftarrow \Big[
    \left(\frac{\partial f_{b,1}(w)}{\partial w}\right)^\top
    \left(\frac{\partial f_{b,3}(w)}{\partial w}\right)^\top \cdots 
    \left(\frac{\partial f_{b,n_b}(w)}{\partial w}\right)^\top
    \Big]$
\STATE $\tilde{G} \leftarrow G - U (U^\top G)$
\STATE $Y \leftarrow [y_{b,1};\; y_{b,2};\; \cdots ; y_{b,n_b}]$
\STATE $F \leftarrow [f_{b,1}(w);\; f_{b,2}(w);\; \cdots; f_{b,n_b}(w)]$
\STATE $w \leftarrow w - \tilde{G}\big(G^\top \tilde{G}\;\big)^{-1}(F-Y)$
\STATEx \(\triangleright\) Update orthonormal basis
\FOR {$g$ in $G$}
\STATE $\tilde{g} \leftarrow g - U (U^\top g)$
\STATE $U \leftarrow [U\; \tilde{g}/\lVert \tilde{g} \rVert]$
\ENDFOR
\ENDFOR
}
\end{algorithmic} 
\end{algorithm}

\begin{algorithm}[!t]
\caption{{\sf Batch-ORFit} with memory limit $m$}
\label{alg:borfit_ipca}
\begin{algorithmic}[1]
\renewcommand{\algorithmicrequire}{\textbf{Input:}}
\renewcommand{\algorithmicensure}{\textbf{Output:}}
\jdiff{
\REQUIRE Data batch sequence $\left(\{(x_{b,k}, y_{b,k})\}_{k=1}^{n_b}\right)_{b=1}^{B}$
\ENSURE  The optimal parameter $w$
\STATE \textbf{Initialize} $U\leftarrow [\;\;],\; \Sigma\leftarrow[\;\;], \; w\leftarrow w_0$
\FOR {$b = 1,2,\dots,B$}
\STATEx \(\triangleright\) Update parameter
\STATE $G \leftarrow \Big[
    \left(\frac{\partial f_{b,1}(w)}{\partial w}\right)^\top
    \left(\frac{\partial f_{b,3}(w)}{\partial w}\right)^\top \cdots 
    \left(\frac{\partial f_{b,n_b}(w)}{\partial w}\right)^\top
    \Big]$
\STATE $\tilde{G} \leftarrow G - U (U^\top G)$
\STATE $Y \leftarrow [y_{b,1};\; y_{b,2};\; \cdots ; y_{b,n_b}]$
\STATE $F \leftarrow [f_{b,1}(w);\; f_{b,2}(w);\; \cdots; f_{b,n_b}(w)]$
\STATE $w \leftarrow w - \tilde{G}\big(G^\top \tilde{G}\;\big)^{-1}(F-Y)$
\STATEx \(\triangleright\) Update orthonormal basis
\STATE $G_{\mathrm{orth}} \leftarrow \mathrm{Orthogonalize } \;\tilde{G}$
% \STATE $R \leftarrow \begin{bmatrix} \Sigma & U^\top  \tilde{g}\\ 0 & u^\top \tilde{g}\end{bmatrix}$
\STATE $\Tilde{U}, \Sigma \leftarrow$ Compute SVD of $\begin{bmatrix} \Sigma & U^\top G\\ 0 & G_{\mathrm{orth}}^\top \tilde{G}\end{bmatrix}$ % = \Tilde{U}\Tilde{\Sigma}\Tilde{V}^\top
% \STATE $\Sigma \leftarrow \Tilde{\Sigma}$
\STATE $U \leftarrow [U\; G_{\mathrm{orth}}]\Tilde{U}$
\STATE $U, \Sigma \leftarrow$ top $m$ singular vectors/values in $U, \Sigma$
\ENDFOR
}
\end{algorithmic} 
\end{algorithm}

\section{Theoretical Results} \label{sec:theory}
In this section, we provide the theoretical properties of the proposed method.
We begin by discussing a formal connection between the proposed method and RLS.

\subsection{Connection to RLS} \label{subsec:connection}

It turns out {\sf ORFit} \jdiff{applied to a linear model $f(x;w) = x^\top w \in\mathbb{R}$} corresponds to an extreme case of the well-known RLS method, as formally described in the following result; see Appendix~\ref{app:proof_compare_RLS} for a proof.

\begin{proposition} \label{prop:compare_RLS}
Consider a linear overparameterized ($p\geq K$) model \jdiff{$f(x;w) = x^\top w \in\mathbb{R}$ and a data sequence $\left((x_k, y_k)\right)_{k=1}^{K}$}. Let $w_0$ be the initialization and $m$ be the memory limit for {\sf ORFit}. Then, at each iteration $i\leq m$, the update rule of \mbox{{\sf ORFit}} results in the same parameter vector as the EW-RLS update rule~\eqref{eq:rls_update} does with $\lambda=0$, $\Pi=I$, and initialization $w_0$.
In this setting, $P_i$ in~\eqref{eq:rls_update} is the projection matrix onto the subspace orthogonal to $\mathrm{span}\{\nabla f_k(w_{k-1})\}_{k=1}^{i}$. 
\end{proposition}
\begin{remark}
    The optimization problem of EW-RLS~\eqref{eq:weighted_ls} is not well-defined for $\lambda=0$. That said, the update rule~\eqref{eq:rls_update} can be still computed for $\lambda=0$, and {\sf ORFit} finds the same solution as the limiting case of EW-RLS.
\end{remark}
\begin{remark}
    {\sf ORFit} in~\eqref{eq:orfit1_infmem} has $O(ip)$ time and memory complexities, compared to those of $O(p^2)$ for the EW-RLS update rule, where typically $i\ll p$ in the overparameterized setting.
\end{remark}

One notable aspect of {\sf ORFit} is that it bridges the two seemingly distinct algorithms OGD and RLS through Proposition~\ref{prop:compare_RLS}. The connection provides us new insights into understanding the behavior of our proposed method, as we discuss next.

\subsection{Characterizing the Solution of {\sf ORFit}}
\label{subsec:solution}

Before presenting our main result, we first provide some intuitions.
To understand the behavior of {\sf ORFit}, let us first recall that the EW-RLS update rule~\eqref{eq:rls_update} is the solution to the optimization problem~\eqref{eq:weighted_ls}.
In light of Proposition~\ref{prop:compare_RLS}, one might be tempted to claim that {\sf ORFit} in~\eqref{eq:orfit1_infmem} solves \eqref{eq:weighted_ls} with $\lambda=0$ and $\Pi=I$.
However, \eqref{eq:weighted_ls} is not well-defined for $\lambda=0$.

Nevertheless, intuitively one can regard {\sf ORFit} as solving \eqref{eq:weighted_ls} in the limit of $\lambda \to 0^+$. 
Then, for sufficiently small $\lambda>0$, the first term in the objective of \eqref{eq:weighted_ls} outweighs the second term, which suggests that, in the overparameterized case, the solution should enforce $y_k \approx w^\top x_k$ for all $k=1,2,\dots, i$, while minimizing $\lVert w-w_0 \rVert^2$.
In the following theorem, we formalize this intuition and characterize the solution of {\sf ORFit}; see Appendix~\ref{app:proof_implicit_bias} for a proof.

\begin{theorem} \label{thm:implicit_bias}
Consider a linear overparameterized ($p\geq K$) model \jdiff{$f(x;w) = \Phi(x)^\top w \in\mathbb{R}^c$ and a data sequence $\left((x_{k}, y_{k})\right)_{k=1}^{K}$}. Let $w_0$ be the initialization and $m$ be the memory limit.
Then, at each iteration $i\leq m$, the parameter vector obtained by {\sf ORFit} is the solution of the following optimization problem:
\begin{equation}
\begin{aligned}
    w_i = \argmin_w&& & \lVert w-w_o \rVert_2\\[-1ex]
    \textup{s.t.}&& & \jdiff{y_{k} = \Phi(x_{k})^\top w \quad(k\in [i]).}
\end{aligned}
\label{eq:implicit_bias}
\end{equation}
\end{theorem}

It is known that, for a linear overparameterized model \jdiff{$f(x;w) = x^\top w$}, in the standard multi-pass learning setting over the dataset $\{(x_k,y_k)\}_{k=1}^{i}$, as the number of iterations goes to infinity, the iterates of stochastic gradient descent (SGD) initialized at $w_0$ with a sufficiently small step size converge to the solution of problem~\eqref{eq:implicit_bias} (see, \emph{e.g.}, Proposition~1 in~\cite{azizan2022stochastic}). Thus, {\sf ORFit} with just an epoch of training finds the solution that SGD in the limit of infinite number of iterations converges to.
\begin{corollary}
Consider a linear overparameterized ($p\geq K$) model \jdiff{$f(x;w) = x^\top w \in\mathbb{R}$ and a data sequence $\left((x_k, y_k)\right)_{k=1}^{K}$}. Let $w_0$ be the initialization and $m$ be the memory limit. The parameter vector obtained by {\sf ORFit} at each iteration $i\leq m$ is equal to what SGD would converge to by iterating over the dataset $\{(x_k, y_k)\}_{k=1}^{i}$ with a sufficiently small step size in the limit of infinite number of iterations.
\end{corollary}

\jdiff{
While Theorem~\ref{thm:implicit_bias} characterizes {\sf ORFit} as solving a global optimization problem, we can also interpret each update step as solving a local optimization problem. This perspective provides an alternative approach to proving Theorem~\ref{thm:implicit_bias} by showing that iteratively solving these local optimization problems ultimately leads to solving the global optimization problem~\eqref{eq:implicit_bias}. This approach is elaborated in detail in~\cite{cho2024pi}.
By examining the update rule~\eqref{eq:orfit2_infmem}, we can recursively characterize each update step as below; see Appendix~\ref{app:proof_local_opt} for a proof.
\begin{proposition}\label{prop:local_opt}
Consider a linear overparameterized ($p\geq K$) model $f(x;w) = \Phi(x)^\top w \in\mathbb{R}^c$ and a data sequence $((x_{k}, y_k))_{k=1}^{K}$. Let $w_0$ be the initialization and $m$ be the memory limit.
Then, at each iteration $i\leq m$, the parameter vector obtained by {\sf ORFit} is the solution of the following optimization problem:
\begin{equation}\label{eq:local_opt}
\begin{aligned}
    w_i = \argmin_{w}&& & \lVert w-w_{i-1} \rVert_2\\[-1ex]
    \textup{s.t.}&& & y_{k} = \Phi(x_k)^\top w \quad(k\in [i]).
\end{aligned}
\end{equation}
\end{proposition}
}

\subsection{\jdiff{Minimax Optimality of the Principal Directions}}
\label{sec:pca_theory}
\jdiff{
{\sf ORFit} utilizes IPCA to summarize the previous gradients with the top principal components to which the new update is orthogonal. We show that this choice is minimax optimal, minimizing the worst-case forgetting for unknown future updates as below. See Appendix~\ref{app:proof_pca_analysis} for a proof.
\begin{proposition}\label{prop:pca_analysis}
Consider a linear overparameterized ($p\geq K$) model $f(x;w) = \Phi(x)^\top w\in\mathbb{R}^c$ and a data sequence $\left((x_k, y_k)\right)_{k=1}^{K}$.
Let $m$ be the memory limit for {\sf ORFit} and $\mathrm{PCA}_{m}\left(\cdot\right)$ denote the span of the top $m$ left singular vectors. Then, at each iteration $i\geq m$, for the accumulated gradients $G_{1:i}:=\begin{bmatrix} \left(\frac{\partial f_{k}(w_{k-1})}{\partial w}\right)^\top\end{bmatrix}_{k=1}^{i}$, $\mathrm{PCA}_{m}\left(G_{1:i}\right)$ is the summary of the memory that minimizes the worst-case forgetting of {\sf ORFit} in the sense that
\begin{equation}
\begin{aligned}
    \mathrm{PCA}_{m}\left(G_{1:i}\right) = \underset{S\in \mathcal{V}_{m}}{\mathrm{arg\, min}} \, \max\limits_{\left\|\Delta w\right\|=1} & \sum_{k=1}^i (f_k(w+\Delta w)-f_k(w))^2\\
    \text{s.t.} \;\;\; & \Delta w \perp S
\end{aligned},
\label{eq:pca_opt}
\end{equation}
where $\mathcal{V}_{m}$ denotes the set of all $m$-dimensional subspaces of $\mathbb{R}^p$.
\end{proposition}
With the limited memory size $m$, {\sf ORFit} stores an approximation $S$ of the accumulated gradients that spans an $m$-dimensional subspace of $\mathbb{R}^p$. Then, the unknown future update $\Delta w$ is constrained to be orthogonal to $S$. Thus, as shown in~\eqref{eq:pca_opt}, the summary through the top $m$ left singular vectors $\mathrm{PCA}_{m}\left(G_{1:i}\right)$ minimizes the worst-case forgetting (\emph{i.e.}, the maximum change of function values among all feasible future update directions $\Delta w$) for the previous datapoints $\left\{(x_k,y_k)\right\}_{k=1}^i$.
}

\section{Experiments}
\label{sec:exp}
%\jdiff{
%\subsection{Rotated MNIST}
%}
In this section, we demonstrate the effectiveness of our proposed methods in the one-pass learning setting and corroborate the theoretical results presented in Section~\ref{sec:theory}. We performed experiments for linear models in the \emph{Rotated MNIST} setup described in~\cite{sharma2021sketching}. In this setup, the inputs are rotated MNIST images for digit `$2$', whose size is $28\times 28$. 
Our goal is to estimate the rotated angles in $[0, \pi]$.
For the training dataset, the angles are uniformly sampled from $[0, \pi]$, and to introduce distribution shift, we order the dataset so that an image rotated with a smaller angle arrives earlier.

\begin{figure}[t]
	\centering
	\begin{subfigure}[b]{.45\textwidth}
	    \centering
	    \includegraphics[width=\textwidth]{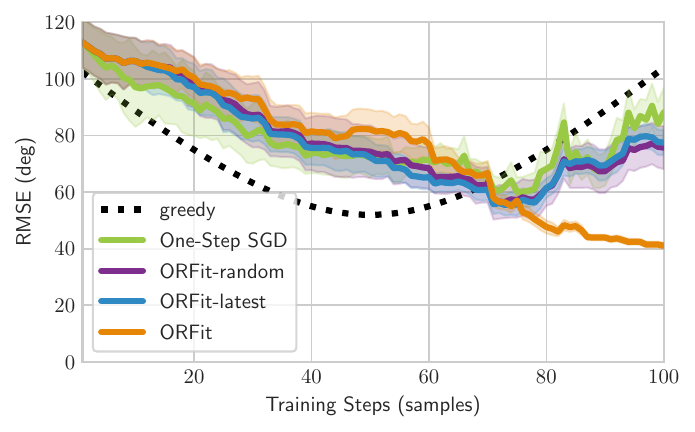}
	    \caption{Test Error}
	    \label{subfig:limited_test}
    \end{subfigure}
    \begin{subfigure}[b]{.45\textwidth}
        \centering
	    \includegraphics[width=\textwidth]{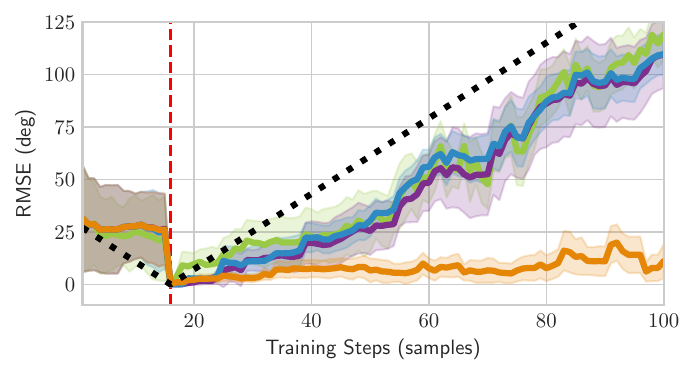}
	    \caption{Sample Prediction Error}
	    \label{subfig:limited_sample}
    \end{subfigure}
	\caption{Results for the memory-restricted setting (\S\ref{subsec:exp_limit}). (\protect\subref{subfig:limited_test}) shows the evolution of the test errors measured after learning each datapoint, while (\protect\subref{subfig:limited_sample}) shows the evolution of the prediction errors for a particular sample (the $16$-th example) after each iteration. The red dashed line indicates the step on which the sample is trained. The shades indicate the standard deviations over $10$ independent runs}
	\label{fig:limited}
\end{figure}

%\jdiff{
\subsection{Learning with Restrictions on Memory Size}
%}
\label{subsec:exp_limit}
In this experiment, we demonstrate the effectiveness of our proposed method in the memory-restricted setting, in which 100 datapoints are sequentially learned while we are only allowed to store up to $10$ basis vectors.
%Note that the size of a single basis vector is the same as that of a single datapoint.
For comparison, we consider the following baselines: (1) \emph{Greedy scheme}: outputs only the label of the most recently learned datapoint, regardless of the input, as an extreme case of forgetting. (2) \emph{One-Step SGD}: employs the one-step learning scheme with the step size in~\eqref{eq:stepsize} but with empty orthogonal basis. (3) \emph{{\sf ORFit}-random}: {\sf ORFit} that keeps $10$ randomly chosen basis vectors after each iteration instead of performing IPCA. (4) \emph{{\sf ORFit}-latest}: {\sf ORFit} that keeps the latest $10$ basis vectors after each iteration instead of performing IPCA.

We first compared the test errors of the proposed method and the baselines. The test errors are measured with the test dataset that consists of $1032$ images of digit `$2$' in the MNIST test set of which each is rotated with a random angle in $[0, \pi]$. As shown in Fig.~\ref{subfig:limited_test}, {\sf ORFit} outperforms other baselines after a sufficient number of training steps. Notably, {\sf ORFit} results in lower variances over the $10$ independent runs with different initialization. 
%For \emph{greedy scheme}, we simply plot the expected error for a sample with its label following the uniform distribution over $[0, \pi]$.
%The plot in the first half of the training shows that the test error can be decreased even with the complete forgetting of previous predictions due to the order of the dataset. Thus, we further look into another metric to demonstrate the effectiveness of the proposed method in terms of forgetting.

Next, we compared the degrees of forgetting for the proposed method and the baselines by keeping track of the prediction errors of a training datapoint throughout the training.
As shown in Fig.~\ref{subfig:limited_sample}, {\sf ORFit} successfully keeps the prediction error low throughout the training.
This is in stark contrast with other methods for which the prediction error quickly increases as other datapoints are learned.
%The result demonstrates that our proposed way of incrementally summarizing the orthogonal basis is much more effective than other methods.
%: by utilizing the structure of the data, {\sf ORFit} extracts important directions where the predictions on the previous data are largely changing.

%\jdiff{
\subsection{Learning without Memory Restriction} %}
\label{subsec:no_memory}
In this experiment, we follow the setup in Section~\ref{subsec:exp_limit} except that this time, we do not impose any memory restrictions. For comparison, we run vanilla SGD with a fixed step size $10^{-5}$. Vanilla SGD makes multiple passes over the entire dataset for $1000$ epochs. %While the first epoch learns the data with the same sequence as the other one-pass learning methods, the entire data is revisited with random orders from the second epoch.

\begin{figure}[t]
	\centering
	\begin{subfigure}[b]{.45\textwidth}
	    \centering
	    \includegraphics[width=\textwidth]{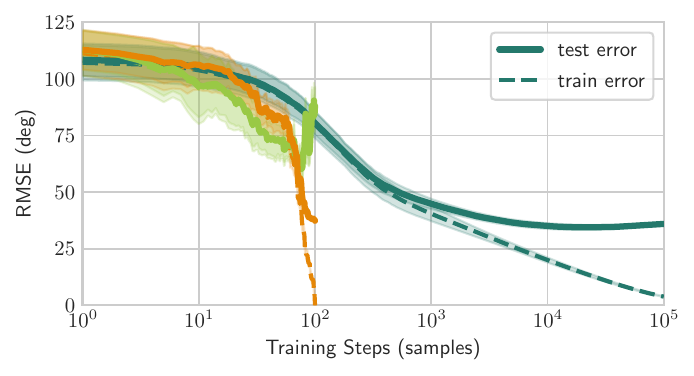}
	    \caption{\jdiff{Training and Test Errors}}
	    \label{subfig:full_test}
    \end{subfigure}
    \begin{subfigure}[b]{.45\textwidth}
        \centering
	    \includegraphics[width=\textwidth]{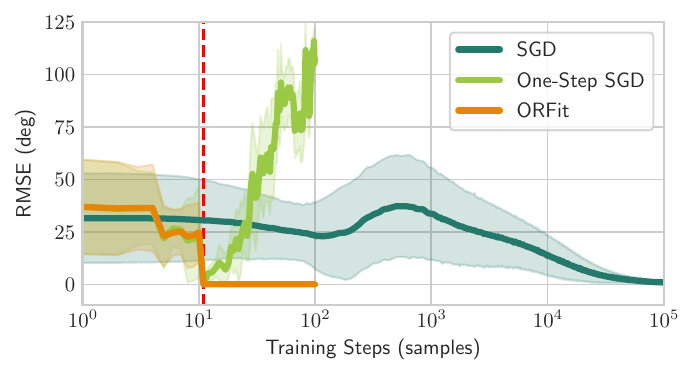}
	    \caption{Sample Prediction Error}
	    \label{subfig:full_sample}
    \end{subfigure}
	\caption{Results for the setting without memory restriction (\S\ref{subsec:no_memory}). (\protect\subref{subfig:full_test}) shows the evolution of the test and train errors measured after each training step, while (\protect\subref{subfig:full_sample}) shows the evolution of the prediction errors for a particular sample (the $11$-th example) after each iteration.
	The red dashed line indicates the step on which the sample is trained.
	The shades indicate the standard deviations over $10$ independent runs.}
	\label{fig:full}
\end{figure}

Reported in Fig.~\ref{subfig:full_test} are the training and test errors of vanilla SGD, One-Step SGD, and {\sf ORFit}. Note that both SGD and {\sf ORFit} learn the training dataset perfectly with almost zero training error. Moreover, after the training is finished, SGD and {\sf ORFit} achieve similar test errors, corroborating Theorem~\ref{thm:implicit_bias}.

In addition, Fig.~\ref{subfig:full_sample} shows the prediction error of the $11$-th training datapoint throughout the training (analogous to Fig.~\ref{subfig:limited_sample}).
Note that {\sf ORFit} perfectly preserves the prediction error of the sample, while One-Step SGD quickly deteriorates it. This result demonstrates the effectiveness of the orthogonal update in {\sf ORFit} for one-pass learning.

\section{Extension to Deep Learning}\label{sec:deep}
 
In this section, we extend our discussion to nonlinear models under the neural tangent kernel (NTK) regime~\cite{jacot2018neural,lee2019wide}.
%by connecting deep neural networks to kernel methods, and it was shown that the NTK remains the same during training the networks in the infinite-width limit through gradient descent. Also, \cite{lee2019wide} showed that a wide neural network is approximated well as a linear model from the first-order Taylor expansion around its initialization with empirical results on finite practically sized networks.
The main idea behind the NTK regime is that when the width of the neural network is chosen large enough, the model is well-approximated  by its first-order approximation around the  initialization:
\begin{equation}
    f_k(w) \approx f_k(w_0) + \nabla f_k(w_0)^\top (w - w_0).
\end{equation}
In particular, Lee et al. \cite{lee2019wide} discussed sufficient conditions (in terms of the width of the network)  for which this approximation is valid; see their Theorem 2.1 for details. Under the linearized regime, we consider expanding the model around the parameter learned at the previous step as
\begin{equation} \label{eq:NTK}
    f_k(w) \approx f_k(w_{k-1}) + \nabla f_k(w_{k-1})^\top (w - w_{k-1})=\vcentcolon f_{k|k-1}(w).
\end{equation}
Throughout, we \jdiff{use $f_{k|k-1}(w)$ to denote} the RHS of~\eqref{eq:NTK}.

A notable feature of {\sf ORFit} is that it is applicable to any differentiable nonlinear model $f$. This is in contrast to the EW-RLS algorithm~\eqref{eq:rls_update}, which is only applicable to linear models $f(x;w)=w^\top x$.
Based on this observation,  we employ the step size calculated in~\eqref{eq:stepsize} to fit the new data for a nonlinear model under the NTK regime~\eqref{eq:NTK}.
More specifically, we consider an analog of the EW-RLS \eqref{eq:weighted_ls} for nonlinear models:
\begin{equation}
    w_i = \arg\min_w \sum_{k=1}^{i} \lambda^{i-k}(y_k - f_{k|k-1}(w))^2 + \lambda^i \lVert w-w_0 \rVert^2_\Pi ,
    \label{eq:weighted_ls_ntk}
\end{equation}
given a forgetting factor $0<\lambda\leq 1$ and $\Pi\succ 0$.
Then similarly to \eqref{eq:rls_update}, one can express the solution of~\eqref{eq:weighted_ls_ntk}  in a recursive manner, while treating $(\nabla f_k(w_{k-1}), y_k-f_k(w_{k-1})+\nabla f_k(w_{k-1})^\top w_{k-1})$ as the streamed data at the $k$-th step:
\begin{equation}
\begin{cases}
    w_i = w_{i-1} + \dfrac{P_{i-1} \nabla f_{i}(w_{i-1}) (y_i - f_{i}(w_{i-1}))}{\lambda^i + \nabla f_{i}(w_{i-1})^\top  P_{i-1} \nabla f_{i}(w_{i-1})}\,, \\
    P_i = P_{i-1} - \dfrac{P_{i-1} \nabla f_{i}(w_{i-1}) \nabla f_{i}(w_{i-1})^\top  P_{i-1}}{\lambda^i + \nabla f_{i}(w_{i-1})^\top P_{i-1} \nabla f_{i}(w_{i-1})}\,,
\end{cases}
\label{eq:rls_update_general}
\end{equation}
with initialization $w_0$ and $P_0 = \Pi^{-1}$. We call this generalized update rule \emph{NTK-RLS}. Following a similar argument as in Section~\ref{sec:theory}, we obtain the following result; see Appendix~\ref{app:proof_implicit_bias_ntk} for a proof.
\begin{theorem} \label{thm:implicit_bias_ntk}
Consider a (nonlinear) overparameterized model \jdiff{$f(x;w)\in\mathbb{R}$ with $p\geq K$ and a data sequence $\left((x_k, y_k)\right)_{k=1}^{K}$}. Let $w_0$ be the initialization and $m$ be the memory limit for {\sf ORFit}.
Then, at each iteration $i\leq m$, the update rule of {\sf ORFit} results in the same parameter vector as the NTK-RLS update rule~\eqref{eq:rls_update_general} does with $\lambda=0$, $\Pi=I$, and initialization $w_0$.
Moreover, at each iteration $i
\leq m$, the parameter vector obtained by {\sf ORFit} is the solution of the following optimization problem:
\begin{equation}
\begin{aligned}
    w_i = \arg\min_w \;\; & \lVert w-w_o \rVert\\
    \textup{s.t.}\;\;\; & y_k = f_{k|k-1}(w) \;\; k=1,2,\dots, i.
\end{aligned}
\label{eq:implicit_bias_ntk}
\end{equation}
\end{theorem}

Note that, under the NTK regime, for an overparameterized model, we have $f_{k|k-1}(w)\approx f(x_k;w)$, and the solution obtained by {\sf ORFit} in one pass is the same as that of SGD in the standard multi-pass setting (as characterized in, \emph{e.g.}, \cite{azizan2022stochastic}).
Compared to \emph{NTK-RLS}~\eqref{eq:rls_update_general}, {\sf ORFit} in~\eqref{eq:orfit1_infmem} greatly reduces both time and memory complexities from $O(p^2)$ to $O(ip)$. 
This is particularly important for $p\gg i$, common to the overparameterized settings (for instance, $p\approx 11\textup{M}$ in ResNet-18, commonly used for the CIFAR-10 dataset, consisting of 50K samples).

\section{Conclusion}

In this paper, we proposed an algorithm called Orthogonal Recursive Fitting ({\sf ORFit}) to tackle one-pass learning. 
We discussed the connection between the proposed method and orthogonal gradient descent (OGD), a practical algorithm in continual learning, as well as the recursive least-squares (RLS), a well-known method from adaptive filtering. 
Through this connection, we explained the advantages of the proposed method and theoretically characterized its behavior. 
Our theoretical findings reveal that {\sf ORFit} attains the same solution as SGD\jdiff{, despite being a single-pass algorithm.}
We validated our method and its theoretical properties through several experiments and discussed its extensions to nonlinear settings, relevant for deep learning.

We conclude with several interesting future directions.
First, although {\sf ORFit} exhibits outstanding performance in the memory-limited setting, some forgetting is still happening. \jdiff{This} is caused by the information loss from summarizing the orthogonal basis via IPCA. 
Theoretically characterizing how much {\sf ORFit} forgets would be of great \jdiff{value}. 
Moreover, one can also come up with other methods to summarize the memory such as matrix sketching\cite{wright2023sketchogd} and compare them with {\sf ORFit}.
Next, we remark that RLS is a special case of the Kalman filter applied on a static system. Based on this connection, another interesting avenue \jdiff{would be} to build on {\sf ORFit} and \jdiff{potentially} devise efficient learning/estimation methods for dynamic systems.
Lastly, exploring the practicality of {\sf ORFit} for deep neural networks based on a comprehensive set of experiments would be of great interest.

\begin{ack}
This work was supported in part by the MIT-IBM Watson AI Lab, MathWorks, the MIT-Amazon Science Hub, and the MIT-Google Program for Computing Innovation. The authors acknowledge the MIT SuperCloud and Lincoln Laboratory Supercomputing Center for providing computing resources that have contributed to the results reported within this paper. The authors also thank Kwangjun Ahn for his valuable input in the early stages of this work.
\end{ack}

\bibliographystyle{plain} 
\bibliography{ref}

\appendix

\section{Connection to EKF}
\jdiff{
{\sf ORFit} also has an interesting connection to the extended Kalman filter (EKF). For a parameterized model $f(\cdot \;;w)\in\mathbb{R}$ with a sequence of data $((x_k,y_k))_{k=1}^K$, we can regard it as a static system with state vector $w$ and formulate a state-space model as
\begin{equation}
\begin{aligned}
    \Dot{w} & = 0,\\
    y_k & = f(x_k; w) + v_k\\
    & = f_k(w) + v_k  \;\;\;\;\; v_k \sim \mathcal{N}(0,R),
\end{aligned}
\end{equation}
where each data point is considered as a measurement of the state $w$ and $v_k$ is a measurement noise model. Then, for the streamed datapoints, estimation over $w$ can be updated sequentially using EKF from the initial estimation $w_0$ as
\begin{equation}
\begin{aligned}
    \hat{w_k} & = \hat{w_{k-1}} + L_k (y_k - f_k(\hat{w_{k-1}})),\\
    L_k & = \dfrac{Q_{k-1} \nabla f_k(\hat{w_{k-1}})} {f_k(\hat{w_{k-1}})^\top Q_{k-1} \nabla f_k(\hat{w_{k-1}}) + R},\\
    Q_k & = [I - L_k f_k(\hat{w_{k-1}})^\top] Q_{k-1}.
\end{aligned}
\end{equation}
When the measurement noise is zero, \emph{i.e.}, $R=0$, the update rule computes the same solution as {\sf ORFit}.
}

\section{Proof of Lemma~\ref{lem:step}} \label{app:proof_step}

Consider $(x,y)\in T_k$ for any $1\leq k \leq i-1$.
\begin{equation}
    f(x;w') = \phi(x)^\top w' = \phi(x)^\top (w-\eta \tilde{g}) = f(x;w) - \eta \phi(x)^\top \tilde{g}.
\end{equation}
Since the orthogonal basis $S$ spans $\nabla_w f(x;w_k)= \phi(x)$, $\phi(x)$ can be represented as $\phi(x) = \sum_{u\in S} \textup{proj}_{u} \big(\phi(x)\big)$. Then,
\begin{align}
    \phi(x)^\top \tilde{g} &= \Big(\sum_{u\in S} \textup{proj}_{u}\big(\phi(x)\big)\Big)^\top \Big(g- \sum_{v\in S} \textup{proj}_{v}(g)\Big)\\
    &= \Big(\sum_{u\in S} \dfrac{uu^\top}{\lVert u \rVert^2}\phi(x)\Big)^\top \Big(I - \sum_{v\in S} \dfrac{vv^\top}{\lVert v \rVert^2}\Big)g\\
    &= \phi(x)^\top \Big(\sum_{u\in S} \dfrac{uu^\top}{\lVert u \rVert^2} - \sum_{u\in S}\sum_{v\in S} \dfrac{uu^\top vv^\top}{\lVert u \rVert^2\lVert v \rVert^2}\Big) g\\
    &= \phi(x)^\top \Big(\sum_{u\in S} \dfrac{uu^\top}{\lVert u \rVert^2} - \sum_{u\in S} \dfrac{uu^\top}{\lVert u \rVert^2}\Big) g = 0
\end{align}
as $u^\top v = 0$ if $u\neq v$. Thus, $f(x;w')=f(x;w)$.
\qed

\section{Proof of Proposition~\ref{prop:compare_RLS}} \label{app:proof_compare_RLS}
In the update rule of {\sf ORFit}~\eqref{eq:orfit1_infmem}, the new basis vector can be represented as
\begin{align}
    v' &= \nabla f_{i}(w_{i-1}) - \sum_{v\in S_{i-1}} \textup{proj}_{v} (\nabla f_{i}(w_{i-1}))\\
    &= (I-\sum_{v\in S_{i-1}}\frac{v v^\top }{\lVert v \rVert^2}) \nabla f_{i}(w_{i-1}) = Q_{i-1} x_i,
\end{align}
where $Q_{i-1} \vcentcolon= I-\sum_{v\in S_{i-1}} v v^\top/\lVert v \rVert^2$. Similarly,
\begin{equation} \label{eq:direction}
    \tilde{g}_{i-1} = Q_{i-1} \nabla \ell_{i}(w_{i-1}) = Q_{i-1} \ell'(y_i, f_i(w_{i-1})) x_i,
\end{equation}
where $\ell'(\cdot,\cdot)$ denotes the derivative of $\ell(\cdot,\cdot)$ with respect to its second argument.

Putting \eqref{eq:direction} into the parameter update in~\eqref{eq:orfit1_infmem},
\begin{equation} \label{eq:osogd_proj1}
    w_{i} = w_{i-1} + \dfrac{Q_{i-1} x_i}{x_i^\top Q_{i-1} x_i} (y_i - x_i^\top w_{i-1}).
\end{equation}
Also, note that $Q_{i-1}$ is symmetric and satisfies $Q_{i-1}^2 = Q_{i-1}$, and it corresponds to the projection matrix onto the orthogonal complement of the subspace $\mathrm{span}\{S_{i-1}\}=\mathrm{span}\{\nabla f_k(w_{k-1})\}_{k=1}^{i-1}$. With these properties, $Q_i$ can be expressed in a recursive form as
\begin{equation} \label{eq:osogd_proj2}
    Q_i = Q_{i-1} - \frac{v' v'^\top }{\lVert v' \rVert^2} = Q_{i-1} - \frac{Q_{i-1} x_i x_i^\top  Q_{i-1} }{x_i^\top  Q_{i-1} x_i}.
\end{equation}
Then, \eqref{eq:osogd_proj1} and \eqref{eq:osogd_proj2} are equivalent to the EW-RLS update rule~\eqref{eq:rls_update} with $\lambda=0$, $\Pi=I$, and initialization $w_0$, while $Q_0=I=P_0$.
\qed

\section{Proof of Theorem~\ref{thm:implicit_bias}} \label{app:proof_implicit_bias}

We prove \eqref{eq:implicit_bias} using KKT conditions. First, we transform the RHS into an equivalent convex problem:
\jdiff{
\begin{equation} \label{eq:convex}
    \arg\min_w \frac{1}{2}\lVert w-w_o \rVert^2 \; \textup{s.t.}\; y_k = \Phi(x_k)^\top w \quad(k\in [i]).
\end{equation}
Let $Y_{1:i}\vcentcolon=[y_1;\cdots;{y_{i}}]\in\mathbb{R}^{ic}$ and $\Phi_{1:i}\vcentcolon=[\Phi(x_1)\;\cdots\;\Phi(x_i)]\in\mathbb{R}^{p\times ic}$. Then, the Lagrangian of~\eqref{eq:convex} is then represented as
\begin{equation} \label{eq:lagrangian}
    \mathcal{L}(w,\lambda)=\frac{1}{2}\lVert w-w_o \rVert^2 + \lambda^\top (Y_{1:i}-\Phi_{1:i}^\top w).
\end{equation}
Since \eqref{eq:convex} is a convex problem, any primal and dual variables of~\eqref{eq:lagrangian} satisfying KKT conditions are primal and dual optimal. The KKT conditions for a pair of primal and dual variables $(w^*,\lambda^*)$ are
\begin{equation} \label{eq:KKT}
\begin{cases}
    w^* - w_0 - \Phi_{1:i}\lambda^* = 0\\
    Y_{1:i} - \Phi_{1:i}^\top w^* = 0.
\end{cases}
\end{equation}

Now, we show that $w_i$ from the {\sf ORFit} update rule~\eqref{eq:orfit2_infmem} satisfies \eqref{eq:KKT} with some $\lambda_i$ so that $w_i$ is the primal optimal solution of~\eqref{eq:convex} which shows~\eqref{eq:implicit_bias}. From~\eqref{eq:orfit2_infmem},
\begin{equation}\label{eq:stacked_update}
    w_i = w_0 - \sum_{k=1}^{i} \tilde{G}_{k-1}\eta_{k-1},
\end{equation}
where $\eta_{k-1}=\left(G_{k-1}^\top \tilde{G}_{k-1}\right)^{-1} (f_k(w_{k-1})-y_k)\in\mathbb{R}^{c}$. For each $k\in[i]$, 
\begin{equation}
    \tilde{G}_{k-1}
    = \Phi(x_k) - \sum_{v\in S_{k-1}} \mathrm{proj}_v (\Phi(x_k)).
\end{equation}
Since each projection vector $v$ is spanned by the columns of $\Phi_{1:k-1}$, each column of $\tilde{G}_{k-1}$ is spanned by the columns of $\Phi_{1:k}$. Then, the term $\tilde{G}_{k-1}\eta_{k-1}$ is also spanned by the columns of $\Phi_{1:k}$. Thus, the summation term in~\eqref{eq:stacked_update} is in total spanned by the columns of $\Phi_{1:i}$, \emph{i.e.}, there exists some $d_i\in\mathbb{R}^{ic}$ such that
\begin{align}
     w_i = w_0 - \sum_{k=1}^{i} \tilde{G}_{k-1}\eta_{k-1}
     = w_0 - \Phi_{1:i} d_i.
\end{align}
}
By letting $\lambda_i\vcentcolon=d_i$, $(w_i, \lambda_i)$ satisfies the first KKT condition. Also, with the update rule~\eqref{eq:orfit2_infmem}, $w_i$ fits all the data $\{(x_k, y_k)\}_{k=1}^{i}$, which implies the second KKT condition. Thus, $w_i$ is the solution of~\eqref{eq:convex} so that \eqref{eq:implicit_bias} holds.
\qed

\section{Proof of Proposition~\ref{prop:local_opt}}
\label{app:proof_local_opt}
\jdiff{
At each iteration $i\leq m$, {\sf ORFit} follows the update rule~\eqref{eq:orfit2_infmem}. With the orthonormal basis matrix $U$ of $S_{i-1}$,
\begin{equation}
    \tilde{G}_{i-1} = (I-UU^\top)G_{i-1} = Q \Phi(x_i),
\end{equation}
where $ Q:=I-UU^\top$ is the projection matrix onto the orthogonal complement of the previous feature space $\mathrm{span}\{S_{i-1}\}=\mathrm{span}\{\cup_{k=1}^{i-1}\mathrm{col}(\Phi(x_k))\}$. Then, we can rewrite the update step as 
\begin{equation}
    \Delta w_i := w_i-w_{i-1} = (\Phi(x_i)^\top Q)^+ (y_i-f_i(w_{i-1}))
\end{equation} with the Moore–Penrose inverse operation $(\cdot)^+$. Also, $\Delta w_i = Q\Delta w_i$ as the projection matrix $Q$ is idempotent, \emph{i.e.} $Q^2=Q$. Since the Moore–Penrose inverse provides the minimum $\ell^2$-norm solutions for under-determined linear systems,
\begin{equation}\label{eq:opt_scalar}
\begin{aligned}
    \Delta w_i
    &= \argmin_{\Delta w} \|\Delta w\|_2 
    \textup{ s.t. }&
    y_i-f_i(w_{i-1}) = \Phi(x_i)^\top Q \Delta w\\
    &= \argmin_{\Delta w} \|\Delta w\|_2 
    \textup{ s.t. }&
    y_i-f_i(w_{i-1}) = \Phi(x_i)^\top Q \Delta w,\\
    & & \Delta w = Q \Delta w\\
    &= \argmin_{\Delta w} \|Q \Delta w\|_2
    \textup{ s.t. }&
    y_i = \Phi(x_i)^\top (w_{i-1}+Q \Delta w),\\
    & & \Delta w = Q \Delta w\\
    &= \argmin_{\Delta w} \|\Delta w\|_2
    \textup{ s.t. }&
    y_i = \Phi(x_i)^\top (w_{i-1}+\Delta w),\\
    & & \Delta w \perp \mathrm{span}\{S_{i-1}\}
\end{aligned}.
\end{equation}
Since the orthogonality condition $\Delta w_i \perp \mathrm{span}\{S_{i-1}\}$ is satisfied if and only if the previous predictions are preserved, \emph{i.e.}, $f_k(w_i)=f_k(w_{i-1})$ for all $k\in[i-1]$,
\eqref{eq:local_opt} holds by induction.
\qed}

\section{Proof of Proposition~\ref{prop:pca_analysis}}
\label{app:proof_pca_analysis}
\jdiff{
We provide an interpretation of {\sf ORFit}'s update direction by establishing the minimax optimality of the principal components in terms of forgetting through the Courant--Fischer--Weyl min-max theorem, which is a standard result in linear algebra related to the Rayleigh quotient and the intersection of subspaces; See, for example, \cite{fischer1905uber, carlson1983minimax, carlson1984generalized, horn2012matrix} for the proof.

\begin{lemma}[Courant-Fischer-Weyl \cite{fischer1905uber}] \label{lem:CFW_minmax}
Let $A = A^\top \in \mathbb{R}^{n \times n}$ is a matrix whose eigenvalues are $\lambda_{1} \leq \lambda_{2}\leq\cdots\leq\lambda_{n}$ and associated orthonormal eigenvectors are $u_{1}, u_{2}, \cdots, u_{n}$. Also, let $\mathcal{V}_{k}$ denotes the set of $k$-dimensional subspaces of $\mathbb{R}^{n}$. Then, for each $1\leq k\leq n$,
\begin{equation} \label{eq:CFW_minmax}
    \begin{aligned}
        \lambda_{k} &= \min\limits_{W\in \mathcal{V}_{k}} ~ \max\limits_{x\in W, \left\|x\right\|=1} x^\top Ax\\
        &= \max\limits_{W\in \mathcal{V}_{n-k+1}} ~ \min\limits_{x\in W, \left\|x\right\|=1} x^\top Ax
    \end{aligned}
\end{equation}
In addition, $W = \mathrm{span}\left\{u_{1}, \cdots, u_{k}\right\}$ achieves the outer minimum for the first expression of  \eqref{eq:CFW_minmax}, and $W = \mathrm{span}\left\{u_{1}, \cdots, u_{k-1}\right\}^{\perp} = \mathrm{span}\left\{u_{k}, \cdots, u_{n}\right\}$ achieves the outer maximum for the second expression of  \eqref{eq:CFW_minmax}.
\end{lemma}

Utilizing this lemma, we can characterize the meaning of approximating the previous gradients with the top principal directions as minimizing the worst-case forgetting for unknown future updates as below.

Consider a linear model $f\left(x;w\right) = \Phi(x)^\top w$. Then, $G_{1:i}:=\begin{bmatrix} \Phi(x_{1}) & \cdots & \Phi(x_{i})\end{bmatrix}\in\mathbb{R}^{p\times ic}$. Let $\Delta w$ denote the parameter update step. 
The change in the function value at a datapoint $x_{i}$ due to parameter update $\Delta w$ can be written as 
\begin{equation} \label{eq:Delta_f_i}
    \Delta f_{i} := f_{i}\left(w+\Delta w\right)-f_{i}\left( w\right)= \Phi(x_{i})^\top\Delta w.
\end{equation}
Concatenating the changes for all datapoints into a vector gives
\begin{equation} \label{eq:Delta_f}
    \Delta F := \begin{bmatrix} \Delta f_{1} ; \cdots ; \Delta f_{i} \end{bmatrix} = G_{1:i}^\top\Delta w.
\end{equation}
Then, we can represent the forgetting as
\begin{equation}
    \sum_{k=1}^i (f_k(w+\Delta w)-f_k(w))^2
    = \left\| \Delta F \right\|_{2}^{2}
    = \Delta w^\top G_{1:i} G_{1:i}^\top \Delta w.
\end{equation}

Meanwhile, Lemma~\ref{lem:CFW_minmax} states that the outer minimum in
\begin{equation} \label{eq:minmax_Delta_f_norm}
    \begin{aligned}
    \lambda_{p-m} 
    &= \min\limits_{W\in \mathcal{V}_{p-m}} ~
    \max\limits_{\Delta w \in W, \left\|\Delta w\right\|=1}
    \left\| \Delta F \right\|_{2}^{2} \\
    &= \min\limits_{W\in \mathcal{V}_{p-m}} ~
    \max\limits_{\Delta w \in W, \left\| \Delta w \right\|=1} 
    \Delta w^\top G_{1:i} G_{1:i}^\top \Delta w
    \end{aligned}
\end{equation}
is achieved with $W^{*} = \mathrm{span}\left\{u_{1}, \cdots, u_{p-m}\right\}$ where $u_{k}$ represents the orthonormal eigenvector of $G_{1:i} G_{1:i}^\top$ associated with the $k$-th least eigenvalue. Since
\begin{equation*}
\mathrm{PCA}_{m}\left(G_{1:i}\right) = \mathrm{span}\left\{u_{p-m+1},\cdots, u_p\right\} = {W^{*}}^{\perp}
\end{equation*}
where $\perp$ denotes the orthogonal complement of a subspace, an update step $\Delta w \in W^{*}$ is orthogonal to the subspace given by $\mathrm{PCA}_{m}\left(G_{1:i}\right)$. Therefore,
\begin{equation}
\begin{aligned}
    \mathrm{PCA}_{m}\left(G_{1:i}\right) = \underset{S\in \mathcal{V}_{m}}{\mathrm{arg\, min}} \, \max\limits_{\left\|\Delta w\right\|=1} & \sum_{k=1}^i (f_k(w+\Delta w)-f_k(w))^2\\
    \text{s.t.} \;\;\; & \Delta w \perp S
\end{aligned}.
\end{equation}
Thus, the update direction satisfying $\Delta w \perp \mathrm{PCA}_{m}\left(G_{1:i}\right)$ minimizes the maximum of the $\ell_{2}$-norm of the function value change spanning all observed datapoints when $\Delta w$ is confined to the $\left(p-m\right)$-dimensional orthogonal subspace of $\mathbb{R}^{p}$.
\qed
}

\section{Proof of Theorem~\ref{thm:implicit_bias_ntk}} \label{app:proof_implicit_bias_ntk}

We first prove the connection between {\sf ORFit} and NTK-RLS. As in the proof of Prop.~\ref{prop:compare_RLS}, the new basis vector and the projected gradient of the loss in the update rule of {\sf ORFit}~\eqref{eq:orfit1_infmem} can be represented as
\begin{align}
    v' &= Q_{i-1} \nabla f_{i}(w_{i-1}), \label{eq:new_bias_vector_ntk}\\
    \tilde{g}_{i-1} &= Q_{i-1} \ell'(y_i, f_i(w_{i-1})) \nabla f_{i}(w_{i-1}), \label{eq:direction_ntk}
\end{align}
where $Q_{i-1} \vcentcolon= I-\sum_{v\in S_{i-1}} v v^\top/\lVert v \rVert^2$.
Putting \eqref{eq:direction_ntk} into the parameter update in~\eqref{eq:orfit1_infmem},
\begin{equation} \label{eq:osogd_proj1_ntk}
    w_{i} = w_{i-1} + \dfrac{Q_{i-1} \nabla f_{i}(w_{i-1})(y_i - f_i(w_{i-1}))}{\nabla f_{i}(w_{i-1})^\top Q_{i-1} \nabla f_{i}(w_{i-1})}.
\end{equation}
Since $Q_{i-1}$ is symmetric and satisfies $Q_{i-1}^2 = Q_{i-1}$, with \eqref{eq:new_bias_vector_ntk}, $Q_i$ can be expressed in a recursive form as
\begin{equation} \label{eq:osogd_proj2_ntk}
    Q_i = Q_{i-1} - \frac{Q_{i-1} \nabla f_{i}(w_{i-1}) \nabla f_{i}(w_{i-1})^\top  Q_{i-1} }{\nabla f_{i}(w_{i-1})^\top  Q_{i-1} \nabla f_{i}(w_{i-1})}.
\end{equation}
Then, \eqref{eq:osogd_proj1_ntk} and \eqref{eq:osogd_proj2_ntk} are equivalent to the NTK-RLS update rule~\eqref{eq:rls_update_general} with $\lambda=0$, $\Pi=I$, and initialization $w_0$, while $Q_0=I=P_0$.

We then prove \eqref{eq:implicit_bias_ntk} using KKT conditions as in Thm.~\ref{thm:implicit_bias}. First, we transform the RHS into an equivalent convex problem:
\begin{equation} \label{eq:convex_ntk}
    \arg\min_w \frac{1}{2}\lVert w-w_o \rVert^2 \; \textup{s.t.}\; y_k = f_{k|k-1}(w) \;\; k=1,\dots,i.
\end{equation}
Let $\tilde{y}_k\vcentcolon=y_k-f_k(w_{k-1})+\nabla f_k(w_{k-1})^\top w_{k-1}$, $\tilde{y}\vcentcolon=[\tilde{y}_1\;\dots\;\tilde{y}_i]^\top\in\mathbb{R}^i$, and $\tilde{X}\vcentcolon=[\nabla f_1(w_{0})\;\dots\;\nabla f_i(w_{i-1})]^\top\in\mathbb{R}^{i\times d}$. Then, the Lagrangian of~\eqref{eq:convex_ntk} is then represented as
\begin{equation} \label{eq:lagrangian_ntk}
    \mathcal{L}(w,\lambda)=\frac{1}{2}\lVert w-w_o \rVert^2 + \lambda^\top (\tilde{y}-\tilde{X}w).
\end{equation}
Since \eqref{eq:convex_ntk} is a convex problem, any primal and dual variables of~\eqref{eq:lagrangian_ntk} satisfying KKT conditions are primal and dual optimal. The KKT conditions for a pair of primal and dual variables $(w^*,\lambda^*)$ are
\begin{equation} \label{eq:KKT_ntk}
\begin{cases}
    w^* - w_0 - \tilde{X}^\top\lambda^* = 0\\
    \tilde{y} - \tilde{X}w^* = 0\,.
\end{cases}
\end{equation}

Now, we show that $w_i$ from the {\sf ORFit} update rule~\eqref{eq:orfit1_infmem} satisfies \eqref{eq:KKT_ntk} with some $\lambda_i$ so that $w_i$ is the primal optimal solution of~\eqref{eq:convex_ntk} which in turn shows~\eqref{eq:implicit_bias_ntk}. From~\eqref{eq:orfit1_infmem}, $w_i = w_0 - \sum_{k=0}^{i-1} \eta_{k} \tilde{g}_{k}$. Then for each $k$,
\begin{align}
    \tilde{g}_k &= \nabla \ell_{k+1}(w_{k}) - \sum_{v\in S_{k}} \textup{proj}_{v} (\nabla \ell_{k+1}(w_{k}))\\
    &= \ell'(y_{k+1}, f_{k+1}(w_{k})) \nabla f_{k+1}(w_{k}) - \sum_{j=1}^{k} c_{k,j} \nabla f_j(w_{j-1})\\
    & = \tilde{X}^\top d_k,
\end{align}
where $d_k\vcentcolon=[-c_{k,1},\dots,-c_{k,k},\ell'(y_{k+1}, f_{k+1}(w_{k})),0,\dots,0]^\top \in\mathbb{R}^i$. Such $c_{k,j}\in\mathbb{R}$ exists since $\mathrm{span}\{S_{k}\} =\mathrm{span}\{\nabla f_j(w_{j-1})\}_{j=1}^{k}$. Then,
\begin{equation}
    w_i = w_0 - \sum_{k=0}^{i-1} \eta_{k} \tilde{X}^\top d_k = w_0 - \tilde{X}^\top \sum_{k=0}^{i-1} \eta_{k} d_k.
\end{equation}
By letting $\lambda_i\vcentcolon=- \sum_{k=0}^{i-1} \eta_{k} d_k$, $(w_i, \lambda_i)$ satisfies the first KKT condition.

For the second KKT condition, we observe that for $k\leq i$,
\begin{align}
    f_{k|k-1}(w_i) &= f_k(w_{k-1}) + \nabla f_k(w_{k-1})^\top (w_i - w_{k-1})\\
    &= f_k(w_{k-1}) - \nabla f_k(w_{k-1})^\top \sum_{j=k-1}^{i-1} \eta_{j} \tilde{g}_{j}\\
    &= f_k(w_{k-1}) - \eta_{k-1} \nabla f_k(w_{k-1})^\top \tilde{g}_{k-1}\label{eq:third}\\
    &= f_k(w_{k-1}) - (f_k(w_{k-1}) - y_k) = y_k, \label{eq:fourth}
\end{align}
where \eqref{eq:third} is satisfied as $\nabla f_k(w_{k-1}) \perp \tilde{g}_{j}$ for $j\geq k$, and the step-size from~\eqref{eq:stepsize} results in~\eqref{eq:fourth}. Then, $w_i$ satisfies the second KKT condition and hence is the solution of~\eqref{eq:convex_ntk}.
\qed

% \jdiff{
% \section{System Identification}
% Consider a system identification problem of which goal is to estimate the system parameter $w_n\in\mathbb{R}^{p_n}$ for a given nominal dynamics model $f(x;w_n)=\Phi_{n}(x)^\top w_n$ for the system state $x\in\mathbb{R}^d$ and feature matrix $\Phi_{n}(x)\in\mathbb{R}^{c\times p_n}$. Specifically, we want to estimate the parameter from the streaming measurements $\{(x_k,y_k)\}_{k=1}^K$ of the system given an initial estimate $\bar{w}_0\in\mathbb{R}^{p_n}$. In practice, there exist some model error and/or  measurement noise which is caused from some unknown dynamics and not purely random. We model this error/noise using some random features $\Phi_r(x)$ s.t.
% \begin{equation}\label{eq:augmented_system}
%     y = \Phi_{n}(x)^\top w_n + \Phi_{r}(x)^\top w_r
%     = \begin{bmatrix}
%         \Phi_{n}(x) \\ \Phi_{r}(x)
%     \end{bmatrix}^\top
%     \begin{bmatrix}
%         w_n \\ w_r
%     \end{bmatrix}
% \end{equation}
% with the augmented parameter $w_r\in\mathbb{R}^{p_r}$.
% Note that even when the nominal model is not overparameterized (\emph{i.e.}, $p_n<K$), we can make the augmented system~\eqref{eq:augmented_system} overparameterized by using large enough $p_r$. Then, we can apply {\sf ORFit} to update the estimate of parameter online along the stream of data with an initial estimate $\begin{bmatrix}
%     \bar{w}_0 \\ 0
% \end{bmatrix}\in\mathbb{R}^p$ where $p=p_n+p_r$.

% \section{Transfer Learning}

% \section{Neural Network Classifier (nonlinear)}
% }

\end{document}